\documentclass{article}

\usepackage[final]{aiplans_2021}

\usepackage[utf8]{inputenc} 
\usepackage[T1]{fontenc}    
\usepackage[colorlinks=true, citecolor=blue, linkcolor=blue, urlcolor=blue]{hyperref}
\usepackage{url}            
\usepackage{booktabs}       
\usepackage{amsfonts}       
\usepackage{nicefrac}       
\usepackage{microtype}      
\usepackage{xcolor}         
\usepackage{amssymb}        
\usepackage[super]{nth}
\usepackage{colortbl}       
\usepackage{multirow}       
\usepackage{graphicx}
\usepackage{booktabs}
\usepackage{subfig}
\usepackage{caption}
\usepackage{listings}
\usepackage{amsmath,amsthm,amssymb,wasysym,bussproofs, csquotes}

\newcommand{\sep}{\ensuremath{\, | \,}}
\newcommand{\lam}{\ensuremath{\lambda}}
\newcommand{\var}[1]{\ensuremath{\mathrm{#1}}}
\newcommand{\ap}[2]{\ensuremath{#1 \, #2}}
\newcommand{\TUnit}{\ensuremath{\mathrm{Unit}}}
\newcommand{\TBool}{\ensuremath{\mathrm{Bool}}}
\newcommand{\TList}[1]{\ensuremath{\mathrm{List} \, #1}}
\newcommand{\TArr}[2]{\ensuremath{#1 \to #2}}
\newcommand{\Unit}{\ensuremath{()}}
\newcommand{\True}{\ensuremath{\mathrm{True}}}
\newcommand{\False}{\ensuremath{\mathrm{False}}}
\newcommand{\ITE}[3]{\ensuremath{\mathrm{if} \, #1 \, \mathrm{then} \, #2 \, \mathrm{else} \, #3}}
\newcommand{\Nil}{\ensuremath{\mathrm{Nil}}}
\newcommand{\Cons}[2]{\ensuremath{\mathrm{Cons} \, #1 \, #2}}
\newcommand{\Foldr}[3]{\ensuremath{\mathrm{Foldr} \, #1 \, #2 \, #3}}

\newcommand{\fstLC}{\nth{1} LC}
\newcommand{\sndLC}{\nth{2} LC}

\title{Towards Neural Functional Program Evaluation}

\author{Torsten Scholak$^{1}$\thanks{Equal contribution.}$\:$, Jonathan Pilault$^{2*}$,  
Joey Velez-Ginorio$^{3}$ \\ 
$^1$ElementAI, a ServiceNow company, $^2$Polytechnique Montreal \& Mila,
$^3$University of Pennsylvania \\
$^1$\texttt{torsten.scholak@servicenow.com, }$^{2}$\texttt{pilaultj@mila.qc}
}

\begin{document}

  \maketitle

  \begin{abstract}
    This paper explores the capabilities of current transformer-based language models for program evaluation of simple functional programming languages. We introduce a new program generation mechanism that allows control over syntactic sugar for semantically equivalent programs. T5 experiments reveal that neural functional program evaluation performs surprisingly well, achieving high $90\%$ exact program match scores for most in-distribution and out-of-distribution tests. Using pretrained T5 weights has significant advantages over random initialization. We present and evaluate on three datasets to study generalization abilities that are specific to functional programs based on: type, function composition, and reduction steps. Code and data are publicly available at \hyperlink{https://github.com/ElementAI/neural-interpreters}{https://github.com/ElementAI/neural-interpreters}.
  \end{abstract}

\section{Introduction}

Neural models originally developed for natural language processing show promising performance for modeling computer programming languages
\citep{chen2021evaluating,austin2021program}.
This has led to a number of interesting applications,
and deep-learning models are now successfully and routinely applied in tools that assist developers in writing and understanding programs and code.
For instance, neural language models can
synthesize \citep{PGL-010,ellis2020dreamcoder},
complete \citep{chen2021evaluating},
and summarize programs \citep{elnaggar2021codetrans},
whether they are written in mainstream languages like Python and Java, or in domain-specific languages like SQL and regex.

A less explored application of neural models is program evaluation \citep{reed2015neural, zaremba2014learning}.
Here, the challenge is to predict the output of a program given its input.
In software engineering, this problem is solved by program interpreters \citep{nystrom2021crafting},
which are implemented as rule-based, non-differentiable systems that take formal program code as input and produce outputs, effects, and/or errors.
While these interpreters must reject programs that are syntactically incorrect or contain semantic errors, neural interpreters
allow the evaluation of incomplete, informally-, or even incorrectly-specified programs.
Developers could use them to predict 
output or effects of programs 
before they are fully written,
which could aid in their very creation.
Previous studies have focused mainly on simple imperative programming languages, e.g., FORTH \citep{bovsnjak2017programming} and subsets of Python \citep{bieber2020learning}.
For these languages, evaluation is complicated by control flow, state, and mutability.
State-of-the-art techniques solve these complications with specialized, complex model architectures,
but results are not as promising as one might hope (\cite{bieber2020learning} report an accuracy of $62.1\%$ on their task and dataset).

This paper studies neural program evaluation for functional programming languages,
which share few of the complications of imperative languages \citep{feser2016differentiable}. Our emphasis is on lambda calculi,
which sit at the core of modern functional programming languages like Scheme and Haskell \citep{pierce2002types}.
We show that program evaluation can be recast such that it is tractable for a language model based on the standard transformer architecture \citep{vaswani2017attention}.
Rather than predicting a program's output given an input,
we train the model to reduce a program to a form that cannot be reduced further.
This generalizes the notion of program evaluation, as it can also be used to model partial application of functions and programs.

Unlike works that use neural programs to strengthen generalization of models \citep{chen2020symbolicgen,nye2020synthesisgen} on another downstream task, we study the generalization of neural program evaluation explicitly. We propose three data splits to measure ``type generalization'',
``function compositional generalization'',
and ``reduction steps generalization''
of our approach.
Furthermore, we investigate the effect of syntax and abstraction on neural program evaluation. We define two lambda calculi that are equally capable of expressing semantically equivalent programs, yet are different in their syntax. We find that the neural program evaluation problem is slightly less tractable for the lambda calculus with the simpler syntax.
Lastly, we compare two evaluation strategies for lambda calculus, lazy evaluation and eager evaluation, and find that they are similarly tractable.

\section{Evaluation of Lambda Calculus}

Our experiments are based on synthetic data for two lambda calculi.
The first lambda calculus (\fstLC{}) is untyped and has three syntactic constructs:
variable, lambda, and application expressions, i.e.,
\begin{align}
    e &= \var{x} \sep{} \lam \var{x}.e \sep{} \ap{e}{e}
\end{align}
The second lambda calculus (\sndLC{}) adds types and additional syntactic constructs:
\begin{align}
    t &= \TUnit{} \sep{} \TBool{} \sep{} \TList{t} \sep{} \TArr{t}{t} \\
    e &= \var{x} \sep{} \lam \var{x}.e \sep{} \ap{e}{e} \sep{} \Unit{} \sep{} \True{} \sep{} \False{} \sep{} \ITE{e}{e}{e} \sep{} \Nil{} \sep{} \Cons{e}{e} \sep{} \Foldr{e}{e}{e}
\end{align}
Terms are generated exclusively from the \sndLC{} using types to guide the generation.
First, the program type is chosen at random from the set of types $t$.
Terms $e$ are then generated according to the type, beginning at the root of the term tree and proceeding down the branches to the leaves.
Specifically, production is recursive and top to bottom (root to leaves). Generation starts with the outermost type constructor, and proceeds by sampling from all compatible term constructors. Production stops when none of the resulting leaf nodes contains hole terms, $e$.
Care is taken to ensure that the generated program does not contain any free variables, is well-typed, and that it terminates without error.
Terms from the \fstLC{} are obtained by transforming terms from the \sndLC{}. We therefore only consider the normalisable subset of the \fstLC{}.
We use Church encoding to represent syntactic constructs from the \sndLC{} that do not appear in the \fstLC{} \citep{pierce2002types}. For example, lists are represented by their right fold:
\begin{align}
    \Nil{} &= \lam \var{c}.\lam \var{n}.\var{n} &
    \Cons{h}{\tau} &= \lam \var{c}.\lam \var{n}.\var{c} \, h \, (\tau \, \var{c} \, \var{n}) &
    \Foldr{f}{e}{l} &= l \, f \, e
\end{align} where $h$ and $\tau$ are head and tail terms, and where $f$, $e$, and $l$ refer to the combining function, the initial value, and the Church-encoded input list of elements, respectively.

We consider two reduction strategies: lazy and eager evaluation.
Lazy evaluation reduces a term to weak head normal form (WHNF), which is a term that is not an application expression. For example, the term $(\lam \var{x}.\var{x}) (\lam \var{y}.\var{y})$ is reduced to the WHNF $\lam \var{y}.\var{y}$, while the term $\lam \var{y}.(\lam \var{x}.\var{x}) \, \var{y}$ is already in WHNF and thus not reduced further.
Eager evaluation not only reduces a term to WHNF, but also reduces all subterms of the term. We call this the ``deep'' reduction and refer to the result as the deep normal form (DNF). The terms $(\lam \var{x}.\var{x}) (\lam \var{y}.\var{y})$ and $\lam \var{y}.(\lam \var{x}.\var{x}) \, \var{y}$ are reduced to the same DNF $\lam \var{y}.\var{y}$.

We define the number of reduction steps of a term to be the number of reductions performed on its root term and all its subterms. Reduction to WHNF is a special case of reduction to DNF, and the number of reduction steps to WHNF is always smaller or equal to the number of reduction steps to DNF. We use the number of reduction steps to denote the degree of evaluation complexity of a term, with more steps translating to higher complexity. This can then be used to evaluate the performance of our neural program evaluation models.

\section{Experiments}

In our experiments, we train a language model based on the encoder-decoder transformer architecture to reduce programs to their normal forms.
Depending on the experiment, the input to the network is a term in either the \fstLC{} or \sndLC{},
and the output is either the WHNF or the DNF of the input term.

Terms are encoded as strings, which are then converted to sequences of tokens with ids that are used as inputs to the network.
The string representation of a term is obtained by pretty-printing the term in Haskell syntax \citep{jones2003haskell}.
Variables are encoded as \verb|x0|, \verb|x1|, \verb|x2|, etc. and lambda expressions are encoded as \verb|\x0 -> e|.
Application expressions are encoded as \verb|e1 e2|. Where necessary, parentheses are used to denote the precedence of application.
Thus, the \fstLC{} term $(\lam \var{x}.\var{x}) (\lam \var{y}.\var{y})$ is encoded as \verb|(\x0 -> x0) (\x1 -> x1)|.
In the \sndLC{}, \Nil{} becomes \verb|[]|, and $\Cons{e1}{e2}$ becomes \verb|e1 : e2|, while lists of one or more elements are pretty-printed using Haskell's built-in list syntax: \verb|[e1, e2, ...]|.
We deviate from the usual Haskell syntax for if-then-else expressions: \ITE{e1}{e2}{e3} is encoded as \verb|ite e1 e2 e3|.
The type of a \sndLC{} term is not encoded explicitly. Hence, both lambda calculi appear to be untyped to the models except that non-normalisable terms do not appear in the data. Our type-driven generation process avoids generating non-normalising terms.
We distinguish evaluation with variable renaming (VR) and evaluation without variable renaming (NVR). In the former case, the variables in the reduced program are freshly generated in the order of their appearance, while in the latter case, the variable names are preserved during reduction, which we expect to be more tractable.

We use a pretrained T5 model \citep{raffel2020exploring} as our base model. The number of parameters of the T5-Small and T5-Large models can be found in the appendix section \ref{sec:model_sizes}.
We fine-tune the model using the Adafactor optimizer \citep{shazeer2018adafactor} with a maximum learning rate of $10^{-4}$ in a linear decay schedule and a batch size of $2048$. Predictions are made using greedy decoding.

A dataset of 1 million unique examples is generated by sampling from the distribution of terms in the \sndLC{}, which are subsequently converted to the \fstLC{}. We thus use the same dataset in experiments with the \fstLC{} and the \sndLC{}. We chose to generate from \sndLC{} so that we could exploit its types for type-driven generation of terms.
Terms that are too long to fit within the model's maximum input ($512$ tokens) and output lengths ($256$ tokens) are discarded.

Performance is evaluated using the average exact string match metric between the normal forms of the predicted terms and the normal forms of the ground-truth terms. For a given example, if all predicted terms match 100\% with ground-truth terms, an exact match is found. The average is taken by dividing the number of exact matches over the total number of examples.

\begin{table}
\parbox[t][][t]{.475\linewidth}{
    \small
    \centering
    \begin{tabular}[b]{c|cc|cc}
        \toprule
        \multirow{3}{*}{Target} & \multicolumn{4}{c}{Exact Match} \\
        &                         \multicolumn{2}{c}{VR} & \multicolumn{2}{c}{NVR}       \\
        &                         \fstLC{} & \sndLC{}    & \fstLC{} & \sndLC{}           \\
        \midrule
        \rowcolor{gray!20}\multicolumn{5}{c}{\it \textbf{T5-Small Pretrained}}           \\
            WHNF & 0.886 & 0.926 & 0.598 & 0.922       \\
            DNF  & 0.698 & 0.920 &  ---  &  ---        \\
        \rowcolor{gray!20}\multicolumn{5}{c}{\it \textbf{T5-Large Pretrained}}           \\
            WHNF & 0.990 & 0.996 & 0.984 & 0.996       \\
            DNF  & 0.988 & 0.994 &  ---  &  ---        \\
        \rowcolor{gray!20}\multicolumn{5}{c}{\it \textbf{T5-Large from Scratch}}           \\
            WHNF & 0.530 & 0.532 & 0.576 & 0.581       \\
            DNF  & 0.533 & 0.536 &  ---  &  ---        \\
        \bottomrule
    \end{tabular}
    \caption{Results on random splits of the dataset.\label{tab:random-splits}}
}%
\hfill%
\parbox[t][][t]{.475\linewidth}{
    \small
    \centering
    \begin{tabular}[b]{c|cc|cc}
        \toprule
        \multirow{3}{*}{Target} & \multicolumn{4}{c}{Exact Match}                        \\
        &                         \multicolumn{2}{c}{VR} & \multicolumn{2}{c}{NVR}       \\
        &                         \fstLC{} & \sndLC{}    & \fstLC{} & \sndLC{}           \\
        \midrule
        \rowcolor{gray!20}\multicolumn{5}{c}{\it \textbf{T5-Small Pretrained}}           \\
            WHNF & 0.824 & 0.870 & 0.502 & 0.796 \\
            DNF  & 0.464 & 0.740 &  ---  &  ---  \\
        \rowcolor{gray!20}\multicolumn{5}{c}{\it \textbf{T5-Large Pretrained}}           \\
            WHNF & 0.992 & 0.986 & 0.970 & 0.968 \\
            DNF  & 0.932 & 0.926 &  ---  &  ---  \\
        \rowcolor{gray!20}\multicolumn{5}{c}{\it \textbf{T5-Large from Scratch}}           \\
            WHNF & 0.519 & 0.508 & 0.561 & 0.552 \\
            DNF  & 0.493 & 0.480 &  ---  &  ---  \\
        \bottomrule
    \end{tabular}
    \caption{Results on the split-by-type dataset.\label{tab:split-by-type}}
}\\
\parbox[t][][t]{.475\linewidth}{
    \small
    \centering
    \begin{tabular}[b]{c|cc|cc}
        \toprule
        \multirow{3}{*}{Target} & \multicolumn{4}{c}{Exact Match}                        \\
        &                         \multicolumn{2}{c}{VR} & \multicolumn{2}{c}{NVR}       \\
        &                         \fstLC{} & \sndLC{}    & \fstLC{} & \sndLC{}           \\
        \midrule
        \rowcolor{gray!20}\multicolumn{5}{c}{\it \textbf{T5-Large Pretrained}}           \\
            WHNF & 0.892 & 0.938 & 0.928 & 0.958 \\
            DNF  & 0.952 & 0.972 &  ---  &  ---  \\
        \bottomrule
    \end{tabular}
    \caption{Results on the compositional dataset.\label{tab:split-by-function-composition}}
}%
\hfill%
\parbox[t][][t]{.475\linewidth}{
    \small
    \centering
    \begin{tabular}[b]{c|cc|cc}
        \toprule
        \multirow{3}{*}{Target} & \multicolumn{4}{c}{Exact Match} \\
        &                         \multicolumn{2}{c}{VR} & \multicolumn{2}{c}{NVR}       \\
        &                         \fstLC{} & \sndLC{}    & \fstLC{} & \sndLC{}           \\
        \midrule
        \rowcolor{gray!20}\multicolumn{5}{c}{\it \textbf{T5-Large Pretrained}}           \\
            WHNF & 0.686 & 0.930 & 0.658 & 0.934 \\
            DNF  & 0.758 & 0.960 &  ---  &  ---  \\
        \bottomrule
    \end{tabular}
    \caption{Results for splits by reduction steps.\label{tab:reduction-step-splits}}
}%
\end{table}

\paragraph{Random Split}

Our findings on uniform random splits are summarized in Table~\ref{tab:random-splits}.
We report the average exact-match results on a held-out dataset of $500$ examples of the best performing models.
Those models were trained on $90000$ examples for up to 100 epochs.
We find that the results for T5-Large are all close to the maximum, and that the results for T5-Small are worse and exhibit more variability:
For T5-Small, the average exact-match is around $0.92$ for all \sndLC{} reduction tasks, while the numbers for the \fstLC{} are much lower. The \fstLC{} results for WHNF-NVR and DNF-VR are particularly weak, below $0.6$ and $0.7$, respectively. This pattern is also seen in the results for T5-Large to a lesser degree. Interestingly, if we do not use pretrained weights but instead use Xavier Normal initialization \citep{pmlr-v9-glorot10a}, performance drops by close to $40$ percentage points\footnote{Additional experiments not reported in Table~\ref{tab:random-splits} show that training T5 from scratch with default Huggingface initialization reduces the performance of the model. We do not observe numbers above $0.35$.} across the board. This result shows that pretraining has significant upside benefits for evaluating neural functional programs. We intuit that Natural Language Understanding skills accumulated during pretraining translates well into neural program execution tasks. We will leave it for future work to verify this intuition more thoroughly.%

\paragraph{Split by Type}
\label{sec:split-by-type}

In this experiment, reported in Table~\ref{tab:split-by-type}, we split the dataset into two parts based on the types of \sndLC{} terms.
The examples are ordered by frequency of type occurrence. The training set contains the most common types representing $80\%$ of the dataset, with the test set containing the remaining $20\%$ comprised of different and less common types.%
\footnote{The six most common types are: $\TBool{}$, $\TUnit$, $\TList{\TBool{}}$, $\TList{\TUnit{}}$, $\TArr{\TUnit{}}{\TBool{}}$, and $\TArr{\TBool{}}{\TUnit{}}$. These types are among those found in the training set, while the types in the test set are much more nested, e.g., $\TArr{(\TArr{\TUnit{}}{\TList{\TList{\TArr{\TBool{}}{\TList{\TUnit{}}}}}})}{\TList{\TList{\TList{\TList{\TUnit{}}}}}}$.}
We subsample the training set to $80000$ examples and the test set to $500$ examples.
Despite the changes, the results are similar to the uniform random split in Table~\ref{tab:random-splits}.
For pretrained models, performance decrease is strongest for the T5-Small model, while the T5-Large model performs almost at the level of the uniform random split, except for a $6$ percentage-point performance drop on the DNF tasks. We attribute the absence of this drop for the WHNF tasks to the fact that reduction to WHNF tends to be concentrated around the root term, which is well covered by the training set. Similarly to previous experiments, we see that pretraining has a larger impact on performance than model size. The performance drops by around $40$ percentage points with training from scratch. In all experiments, we observe that \sndLC{} performs slightly worse ($-9$ percentage points on average) for large models.

\paragraph{Split by Function Composition}
\label{sec:split-by-function-composition}

For this experiment, we create a new evaluation dataset from the training examples used in the first experiment on random splits.
We produce unique examples by composing terms $e1$, $e2$ from the training set using application, $e1 \, e2$.%
\footnote{To improve the dataset's diversity, each term $e1$, $e2$ can only occur up to $3$ times.
Type checking is performed on the new examples to ensure well-typedness.
Input/output maximum lengths are limited to $512$/$256$ tokens.}
The final dataset contains $500$ examples of each lambda calculus and evaluation strategy.
We use the best performing models on the random splits to evaluate the performance of the new dataset.
Our findings, reported in Table~\ref{tab:split-by-function-composition},
show that the performance on the compositional evaluation dataset is lower compared to the random split. The \sndLC{} results are slightly better than the \fstLC{} results, and among the \fstLC{} results, those on the WHNF-VR task are particularly poor.

We also analyzed the results on the new dataset by token length. \fstLC{} input and target programs are both about $2.5$ times as long as their \sndLC{} counterparts. When compensated for this difference, the dependence of the exact-match performance on the target length is about the same for both languages: it stays high and approximately constant for targets below 100 tokens for \fstLC{} and 50 tokens for \sndLC{}, and then falls off linearly with length. See section \ref{sec:length} for details.

\paragraph{Split by Number of Reduction Steps}
\label{sec:reduction-step-splits}

This last experiment splits the data such that the training examples contain the fewest reduction steps and the test set the most reduction steps.
Since larger reduction step counts imply greater complexity, we expect a performance drop compared to the random split.
Our results are presented in Table~\ref{tab:reduction-step-splits}.

Consider the WHNF reduction task.
The \fstLC{} and \sndLC{} have median reduction step counts of $4$ and $3$, respectively.
In order to cover the bulk of the distribution of reduction steps, training examples have therefore up to $6$ reduction steps. Larger counts up to $12$ steps are in the test set.
We subsampled the training and test sets such that the proportion of examples per reduction step count is the same between the two lambda calculi.
We find that the \sndLC{} model performs much better than the \fstLC{} model on the test set: For \fstLC{}, the performance drops by one third compared to the random split, while for \sndLC{} the performance drops by about $7$ percentage points.
For the DNF reduction task, the \fstLC{} and \sndLC{} have median reduction step counts of $6$ and $4$, respectively.
The training examples have up to $8$ reductions steps, and examples with between $9$ and $32$ reduction steps are test examples. Subsampling equivalent to that of the WHNF task was applied.
The results for DNF reduction are similar to those obtained for WHNF. Again, the drop in performance is bigger for the \fstLC{} model than for the \sndLC{} model.

\paragraph{Input and Output Program Length}
\label{sec:length}

\begin{figure}[!htp]%
    \centering
    \hspace*{-1.5em}
    \subfloat[\centering Input Program]{{\includegraphics[width=0.54\linewidth]{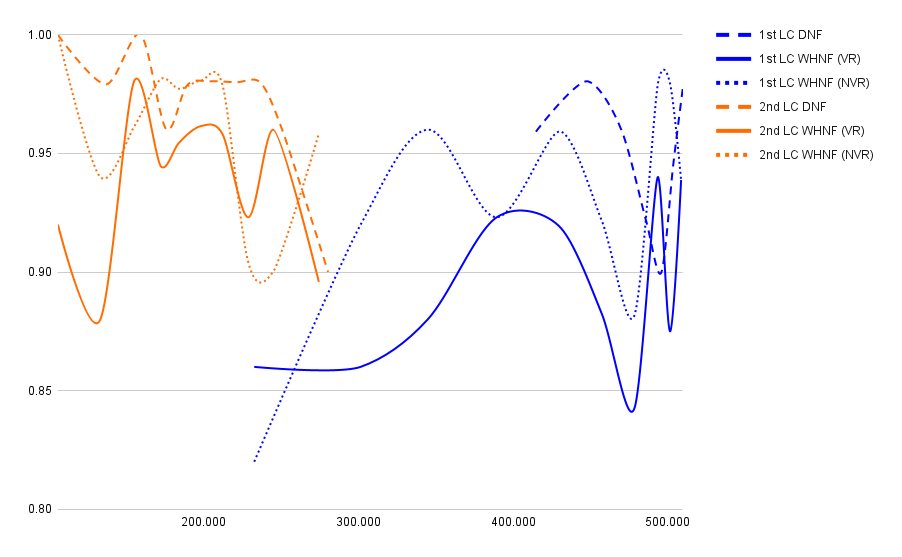} }}%
    \hspace*{-1.0em}
    \subfloat[\centering Output Program]{{\includegraphics[width=0.54\linewidth]{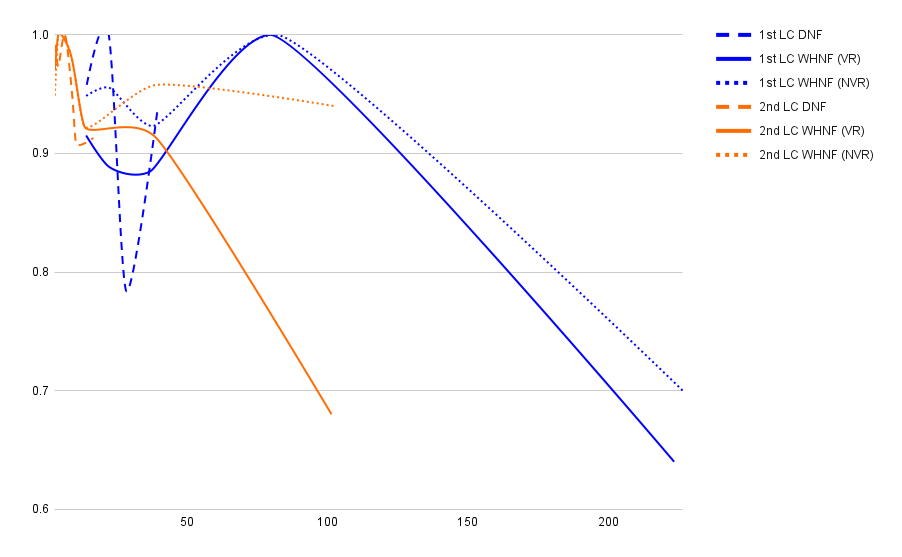} }}%
    \caption{T5-Large Exact Match vs. (a) Input and (b) Output Program Lengths} \label{fig:length} %
\end{figure}

To verify if program input and output lengths (number of tokens) may explain differences between \fstLC{} and \sndLC{}, we show exact match performance for increasing program lengths in Figures \ref{fig:length} (a) and \ref{fig:length} (b). We grouped the data in 10 different program length ranges with equal amounts of examples. The graphs plot the average program length of each range (x-axis) against the average exact match performance (y-axis). \sndLC{} program inputs are generally smaller. However, for WHNF and NVR, when \sndLC{} and \fstLC{} length intersect, \fstLC{} exact match is close to 5 percentage points lower. Interestingly, for \fstLC{}, performance tends to increase with input length as shown in Figure \ref{fig:length} (a). Output program lengths are typically 2.5 times smaller for \sndLC{}. Curves in Figure \ref{fig:length} (b) display  wave patterns with clear downward trends as program length increases. For the same length, the \fstLC{} results are not consistently at or below the \sndLC{} results. Instead, the \fstLC{} results appear shifted and scaled up along the length-axis. Overall, we notice that exact-match performance on the output length is about the same for both languages: it stays high and approximately constant for targets below 50 tokens for \fstLC{} and 100 tokens for \sndLC{} and then falls off linearly with length.

\section{Concluding Remarks}

We compare neural execution of programs in two lambda calculi, \fstLC{} and \sndLC{}. Both express semantically equivalent programs, but \fstLC{} uses a simpler syntax at the cost of longer program length.
Our experiments with T5 have shown near tractable performance for synthetic in-distribution \fstLC{} and \sndLC{} data, but stark performance differences are observed for out-of-distribution data.
We analyze our models' generalization capability on four splits of the data: uniformly random, by program type, by function composition, and by number of reduction steps.
On average, models trained on \fstLC{} generalize less then models trained on \sndLC{}. We observe significant gains from using T5 with pretrained weights compared to randomly initialized models on in-distribution and unseen-type experiments. Input/output program lengths cannot completely account for differences in languages. Our experiments indicate that neural interpretation of functional programs is more tractable when expressed in more ``sugared'' syntax, but further evidence is needed to completely support this claim.
Further, we show that T5-Large is consistently better than T5-Small.
Pretraining on the T5 corpus and tasks yields better results than training from scratch.
Our analysis of different evaluation strategies show that lazy reduction is slightly easier to learn than eager reduction, but may generalize less on nontrivial splits of the data. Results with and without variable renaming do not show a clear trend. This is unexpected since variable renaming is an additional syntactic operation that complicates the task and is not strictly necessary.

Future work on more difficult tasks and data splits will allow us to better understand the generalization properties of our models as well as the impact of syntactic complexity on performance.

\clearpage

\bibliographystyle{ACM-Reference-Format}
\bibliography{bibliography.bib}

\newpage
\appendix

\section{Model sizes} \label{sec:model_sizes}

Model sizes are listed in Table~\ref{table:params}.

\begin{table}[h]
    \centering
    \begin{tabular}{c|c}
    \toprule
    Model & Params (M)  \\
    \midrule
    T5-Small Pretrained & 60 \\
    T5-Small from Scratch & 60 \\
    T5-Large Pretrained & 770 \\
    T5-Large from Scratch & 770 \\
    \bottomrule
    \end{tabular}
    \caption{Number of parameters in millions.}
    \label{table:params}
\end{table}

\section{Example Data} \label{sec:example_data}

The Table~\ref{table:example_data} lists parallel examples of \fstLC{} and \sndLC{} terms and their deep normal forms from our dataset.

\begin{table}[h]
\centering
\begin{tabular}{c|p{12cm}}
\toprule
\fstLC{} &
\begin{lstlisting}
(\x0 -> \x1 -> \x2 -> x1) ((\x3 -> \x4 -> (\x5 -> \x6 -> x6) (\x7 -> (\x8 -> \x9 -> x7) x4) ((\x10 -> \x11 -> \x12 -> \x13 -> x12 x10 (x11 x12 x13)) (\x14 -> x14) (\x15 -> \x16 -> x16))) (\x17 -> \x18 -> x17))
\end{lstlisting} \\
\fstLC{} DNF &
\begin{lstlisting}
\x0 -> \x1 -> x0
\end{lstlisting} \\
\sndLC{} &
\begin{lstlisting}
(\x0 -> True) ((\x1 -> \x2 -> foldr (\x3 -> (\x4 -> \x5 -> x3) x2) [()] []) True)
\end{lstlisting} \\
\sndLC{} DNF &
\begin{lstlisting}
True
\end{lstlisting} \\
\midrule
\fstLC{} &
\begin{lstlisting}
(\x0 -> \x1 -> x1) (\x2 -> \x3 -> \x4 -> \x5 -> \x6 -> x6) (\x7 -> \x8 -> (\x9 -> \x10 -> \x11 -> \x12 -> \x13 -> x13) x8 x7)
\end{lstlisting} \\
\fstLC{} DNF &
\begin{lstlisting}
\x0 -> \x1 -> \x2 -> \x3 -> \x4 -> x4
\end{lstlisting} \\
\sndLC{} &
\begin{lstlisting}
ite False (\x0 -> \x1 -> \x2 -> []) (\x3 -> \x4 -> (\x5 -> \x6 -> \x7 -> []) x4 x3)
\end{lstlisting} \\
\sndLC{} DNF &
\begin{lstlisting}
\x0 -> \x1 -> \x2 -> []
\end{lstlisting} \\
\midrule
\fstLC{} &
\begin{lstlisting}
(\x0 -> \x1 -> x1) (\x2 -> \x3 -> x3) (\x4 -> x4) ((\x5 -> \x6 -> \x7 -> \x8 -> x7 x5 (x6 x7 x8)) ((\x9 -> \x10 -> x10) (\x11 -> x11) (\x12 -> x12)) ((\x13 -> \x14 -> \x15 -> \x16 -> x15 x13 (x14 x15 x16)) ((\x17 -> \x18 -> x18) (\x19 -> x19) (\x20 -> x20) (\x21 -> x21)) ((\x22 -> \x23 -> \x24 -> \x25 -> x24 x22 (x23 x24 x25)) (\x26 -> \x27 -> x26) (\x28 -> \x29 -> x29) (\x30 -> (\x31 -> \x32 -> x32) x30) ((\x33 -> \x34 -> \x35 -> \x36 -> x35 x33 (x34 x35 x36)) (\x37 -> x37) (\x38 -> \x39 -> x39)))) (\x40 -> \x41 -> x41) (\x42 -> x42))
\end{lstlisting} \\
\fstLC{} DNF &
\begin{lstlisting}
\x0 -> x0
\end{lstlisting} \\
\sndLC{} &
\begin{lstlisting}
foldr (\x0 -> \x1 -> x1) (\x2 -> x2) [] (foldr (\x3 -> \x4 -> x4) () ((:) (ite False () ()) ((:) (ite False (\x5 -> x5) (\x6 -> x6) ()) (foldr (\x7 -> (\x8 -> \x9 -> x9) x7) [()] [True]))))
\end{lstlisting} \\
\sndLC{} DNF &
\begin{lstlisting}
()
\end{lstlisting} \\
\midrule
\fstLC{} &
\begin{lstlisting}
(\x0 -> \x1 -> \x2 -> \x3 -> x2 x0 (x1 x2 x3)) ((\x4 -> x4) (\x5 -> x5)) ((\x6 -> \x7 -> \x8 -> x8) ((\x9 -> (\x10 -> \x11 -> \x12 -> \x13 -> x12 x10 (x11 x12 x13)) ((\x14 -> \x15 -> x15) (\x16 -> \x17 -> x17) (\x18 -> \x19 -> x18)) ((\x20 -> \x21 -> \x22 -> x22) x9) (\x23 -> (\x24 -> \x25 -> x24) (\x26 -> x23) (\x27 -> x23)) (\x28 -> \x29 -> x28)) (\x30 -> x30)))
\end{lstlisting} \\
\fstLC{} DNF &
\begin{lstlisting}
\x0 -> \x1 -> x0 (\x2 -> x2) x1
\end{lstlisting} \\
\sndLC{} &
\begin{lstlisting}
(:) ((\x0 -> x0) ()) ((\x1 -> []) ((\x2 -> foldr (\x3 -> ite True (\x4 -> x3) (\x5 -> x3)) True ((:) (ite False False True) ((\x6 -> []) x2))) ()))
\end{lstlisting} \\
\sndLC{} DNF &
\begin{lstlisting}
[()]
\end{lstlisting} \\
\midrule
\fstLC{} &
\begin{lstlisting}
(\x0 -> \x1 -> x1) (\x2 -> (\x3 -> \x4 -> x4) (\x5 -> (\x6 -> (\x7 -> (\x8 -> \x9 -> x9) x7) ((\x10 -> \x11 -> \x12 -> x11) x2)) x2) (\x13 -> x13)) (\x14 -> (\x15 -> \x16 -> x16) (\x17 -> (\x18 -> \x19 -> \x20 -> x20) (\x21 -> x21) (\x22 -> x22)) ((\x23 -> \x24 -> \x25 -> \x26 -> x25 x23 (x24 x25 x26)) ((\x27 -> \x28 -> x28) (\x29 -> \x30 -> x30) (\x31 -> \x32 -> \x33 -> \x34 -> x34) x14) (\x35 -> \x36 -> x36)))
\end{lstlisting} \\
\fstLC{} DNF &
\begin{lstlisting}
\x0 -> \x1 -> \x2 -> x1 (\x3 -> \x4 -> \x5 -> x5) x2
\end{lstlisting} \\
\sndLC{} &
\begin{lstlisting}
foldr (\x0 -> foldr (\x1 -> (\x2 -> (\x3 -> (\x4 -> \x5 -> x5) x3) ((\x6 -> True) x0)) x0) (\x7 -> x7) []) (\x8 -> foldr (\x9 -> (\x10 -> \x11 -> \x12 -> x12) () ()) [foldr (\x13 -> \x14 -> x14) (\x15 -> \x16 -> \x17 -> ()) [] x8] []) []
\end{lstlisting} \\
\sndLC{} DNF &
\begin{lstlisting}
\x0 -> [\x1 -> \x2 -> ()]
\end{lstlisting} \\
\bottomrule
\end{tabular}
\caption{Five program examples. Shown are the equivalent \fstLC{} and \sndLC{} terms and their respective deep normal forms.
\label{table:example_data}}
\label{tab:my_label}
\end{table}

\end{document}